\documentclass[11pt,a4paper]{article}
\usepackage[hyperref]{emnlp2020}
\usepackage{times}
\usepackage{latexsym}
\usepackage{kotex}

\usepackage{microtype}

\aclfinalcopy 
\def\aclpaperid{1986}

\usepackage{tikz}

\usepackage{multirow}
            
\title{Revisiting Modularized Multilingual NMT to Meet Industrial Demands}

\author{Sungwon Lyu$^{1}$, Bokyung Son$^{1,2}$, Kichang Yang$^{1,3}$, Jaekyoung Bae$^{1}$\\
  $^1$Kakao Enterprise / Seoul, Republic of Korea \\
  $^2$Department of Linguistics, Seoul National University \\
  $^3$School of Software, Soongsil University \\
  \texttt{\{james.ryu, meta.mon, kevin.y, storm.b\}@kakaoenterprise.com} \\}

\date{\today{}}

\begin{document}
\maketitle
\begin{abstract}

The complete sharing of parameters for multilingual translation (1-1) has been the mainstream approach in current research. 
However, degraded performance due to the capacity bottleneck and low maintainability hinders its extensive adoption in industries.
In this study, we revisit the multilingual neural machine translation model that only share modules among the same languages (M2) as a practical alternative to 1-1 to satisfy industrial requirements.
Through comprehensive experiments, we identify the benefits of multi-way training and demonstrate that the M2 can enjoy these benefits without suffering from the capacity bottleneck.
Furthermore, the interlingual space of the M2 allows convenient modification of the model.
By leveraging trained modules, we find that incrementally added modules exhibit better performance than singly trained models. 
The zero-shot performance of the added modules is even comparable to supervised models. 
Our findings suggest that the M2 can be a competent candidate for multilingual translation in industries.

\end{abstract} 

\begin{figure*}[h]
  \centering
  \includegraphics[width=\textwidth]{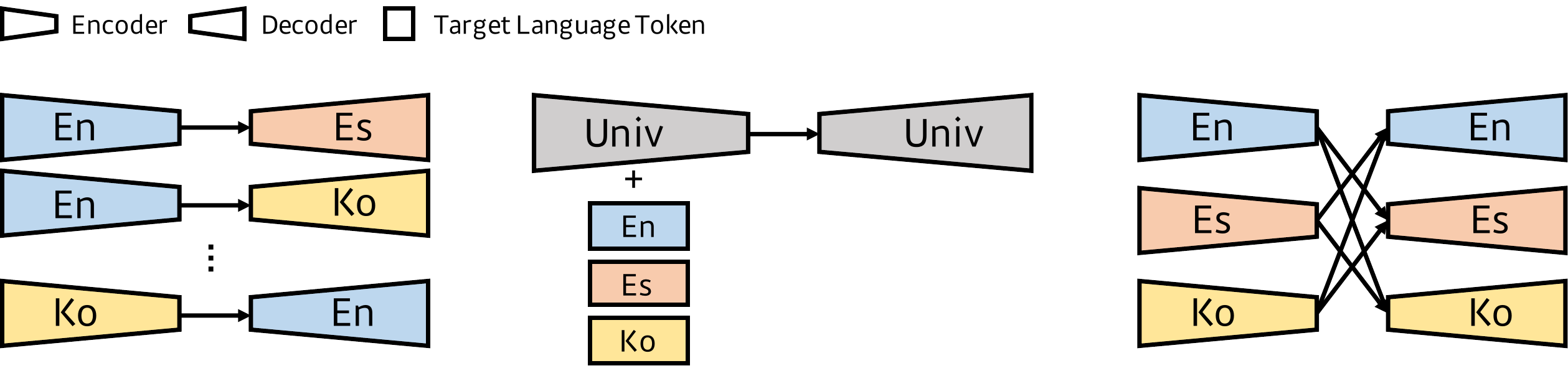}
  \caption{\textbf{Model overview} of three different types of multilingual NMT models for three languages: En, Es, Ko. \textit{Left} is a collection of single models for 6 translation directions. \textit{Middle} is the 1-1 model that share the whole parameters of the model for 6 directions. \textit{Right} is the M2 model that only share language-specific modules.}
  \label{fig:models}
\end{figure*}

\section{Introduction}

With the current increase in the demand for neural machine translation (NMT), serving an increasing number of languages poses a practical problem for the industry.
A naive approach for multilingual NMT is to have multiple single-directional models, which is unsustainable owing to the quadratic increase of models as more languages are introduced. 
A more practical approach is to limit the number of models by sharing the components among the models \citep{dong2015multi, firat2016multi, ha2016toward, johnson2017google}. 
In addition to reducing the number of parameters, sharing the components is also regarded as an effective method to enhance the performance.
A fully shared model (henceforth 1-1), which only uses one encoder and one decoder to translate all directions \citep{ha2016toward,johnson2017google}, has been the most popular method because of its compactness.

However, introduction of a significant number of tasks into a 1-1 model is known to cause capacity bottleneck.
\citet{aharoni2019massively} suggested that, given a fixed model capacity, a 1-1 model is bound to the tradeoff between the number of languages and translation accuracy.
\citet{zhang2020improving} explicitly identified the capacity bottleneck problem of the 1-1 model by showing a clear decrease in performance when translation directions are doubled.
Moreover, data unbalance complicates the problem.
\citet{arivazhagan2019massively} presented the transfer and interference dilemma among low and high resource languages in an unbalanced environment. 

The capacity bottleneck observed in the 1-1 model is particularly undesirable for the industry.
Unlimited scaling of the model size \citep{zhang2020improving} is impossible in practice, where inference cost and latency are crucial. 
With limited capacity, gain from multilingual translation training (henceforth multi-way training) without being subject to the losses of the capacity bottleneck is difficult to achieve.
Furthermore, modification of the 1-1 model such as simple addition of a language is troublesome because the entire model must be retrained from the beginning as a single module, thus requiring a considerable amount of time and effort.
This low maintainability makes 1-1 less attractable for industrial use. 
Still, the benefit from multi-way training is difficult to miss.

These problems lead us to revisit the multilingual neural machine translation model that share parameters among the same languages \citep{firat2016multi}. 
We named this architecture as the \textit{modularized multilingual NMT model} (henthforth M2) since the model share language-specific modules (encoders or decoders) instead of the whole model.
Figure \ref{fig:models} illustrates the architectural overview of multilingual translation using single models, the 1-1 and the M2. 
Although the M2 has not been given substantial attention owing to the linear increase in its parameters as the number of languages increases, it is relatively free from the capacity bottleneck problem while maintaining a reasonable inference cost.
In this study, we explore the possibility of M2 as an alternative to the 1-1 model in industrial settings.

To resolve the capacity bottleneck problem while enjoying the benefits, we identify the effects of multi-way training in a carefully controlled environment.
We find that the \textit{data-diversification} and \textit{regularization} of multi-way training enable the M2 to outperform both single and 1-1 models with less suffering from capacity bottlenecks.
Additionally, the M2 demonstrates a comparable performance increase to 1-1 for low resource pairs in an unbalanced environment.

Combined with its modularizable architecture, interlingual space learned by the M2 allows convenient and effective modification of the model. 
The simple addition of language-specific modules to the M2 outperformed an individually trained model.
The zero-shot learning of the incremented language module outperforms English pivoted translation and is even comparable to a supervised model.
Finally, we show that the language invariance of such space improves with more languages.

In summary, our contribution is threefold. 
1) We conceptually specify the effects of multi-way training and verified them with comprehensive experiments. 
2) We show that the M2 can leverage those effects as the 1-1 without the constraint of the capacity bottleneck. 
3) Finally, we find that multi-way training of the M2 forms interlingual space which allows simple yet effective extension of languages. 

\section{Related works}

\subsection{Neural machine translation}

The most popular framework for NMT is the encoder-decoder model \citep{cho2014learning,sutskever2014sequence,bahdanau2014neural,luong2015effective,vaswani2017attention}. 
Adopting attention module greatly improved the performance of encoder-decoder model by using context vector instead of fixed length vector \citep{bahdanau2014neural,luong2015effective}.
By exploiting multiple attentive heads, the Transformer model has become the de-facto standard model in NMT \citep{vaswani2017attention,ott2018scaling,so2019evolved}. 

\subsection{Multilingual neural machine translation}

\citet{dabre2019survey} categorized the architectures of multilingual NMTs according to their degrees of parameter sharing. 
We briefly introduce the models under their criteria.

Early multilingual NMT models \textit{minimally shared} the parameters by sharing language-specific encoder \citep{dong2015multi,lee2017fully} or decoder \citep{zoph2016multi}.
\citet{firat2016multi} extended this to sharing both language-specific encoders and decoders with a shared attention module.

The 1-1 model, \textit{fully shared}, uses only one encoder and decoder to translate all directions \citep{ha2016toward,johnson2017google}. 
The target language is indicated by prepending a reserved target language token to the source text.
Being compact, the 1-1 model has become the mainstream of multilingual NMT research \citep{ha2016toward,johnson2017google,aharoni2019massively,arivazhagan2019massively,wang2019multilingual,liu2020multilingual}, 
However, subsequent studies tried to solve the capacity bottleneck problem of the 1-1 through knowledge compression \citep{tan2019multilingual}, language clustering \citep{tan2019clustering} or increased capacity \citep{zhang2020improving}.

\textit{Partially shared} models are extensively studied to compromise the capacity bottleneck and model size \citep{blackwood2018multilingual,sachan2018parameter,platanios2018contextual,zaremoodi2018adaptive,bapna2019simple}.
Despite their popularity, we do not compare them in this work because \textit{partially sharing} is essentially relaxing the capacity constraint of \textit{fully sharing}.
Also, \citet{sachan2018parameter} reported that the performance of \textit{partially shared} models is language-specific, which is not the focus of our study. 
Instead, we focus on the general trade off of parameter sharing. 

\subsection{Interlingual representation}

Building interlingual\footnote{We prefer the term `interlingual' to `language-agnostic' because we expect it to be better if the space is shared while maintaining the language-specific features instead of removing them.}  representation is another interest in multilingual language modeling \citep{schwenk2017learning}. 
Interlingual space is the ground for zero-shot translation \citep{johnson2017google,arivazhagan2019missing,al2019consistency} and incremental training \citep{escolano2019bilingual}.
Several explicit methods were suggested to build interlingual space including shared attention \citep{firat2016multi}, neural interlingua module \citep{lu2018neural}, attention bridge \citep{vazquez2019multilingual}, auxiliary loss \citep{arivazhagan2019missing} and shared encoder \citep{sen2019multilingual}.

\bigbreak

We further extend the study of \citet{firat2016multi} which inspired our M2.
\citet{firat2016multi} only shared English encoder and decoder as they used English-centered data (parallel corpus that include English).
Instead we show that sharing modules of all languages using diverse directions of data further increases the performance and is the key to build interlingual representation without any explicit regularization. 

Our motivation to rediscover the M2 is concurrently shared with \citet{escolano2020multilingual}.
\citet{escolano2020multilingual} empirically show that M2 is capable of quickly deploying new languages with incrementally added modules, and found it outperforms 1-1. 
We also experiment on incremental learning and get a similar conclusion, and further interpret the results as an indication that M2 effectively forms an interlingual space. 
Regarding comparison of M2 and 1-1 in general, we deliver an in-depth understanding of a less-studied model M2 focusing on how to maximize its utility in industry. 
Experiments on incremental learning are to check whether M2 is a maintainable alternative to 1-1 (which requires expensive re-training from scratch). 


\section{Effects of multi-way training}
\label{sec:effects}

Because of its complexity, the effects of multi-way training are yet to be identified.
Various factors may affect the performance of multilingual translation: model size compared to the amount of data, the number of training directions, the degree of data imbalance among different directions, and the portion of multi-parallel data.
In this section, we discuss the possible effects on performance resulting from these factors.

\paragraph{Capacity bottleneck}
A capacity bottleneck is the most plausible cause of performance degradation in multi-way training.
For a fixed size model, the capacity bottleneck is more prominent with the increase in training directions (especially target languages) and the amount of data \citep{johnson2017google, aharoni2019massively, arivazhagan2019massively, zhang2020improving}.

\paragraph{Cross-language effect}

Cross-language effect occurs when multiple languages are shared in a module. 
Low resource languages reportedly benefit from multi-way training when trained along with high resource pairs \citep{zoph2016transfer, nguyen2017transfer, neubig2018rapid}.
The interaction among languages in a module can either be positive (transfer) or negative (interference) on the performance according to their similarity in linguistic patterns.

\paragraph{Data-diversification}

Data-diversification is associated with the portion of multi-parallel among multi-way data.
If either the source-side or the target-side language is shared across two directions and data of the directions is not multi-parallel, the shared module learns more diverse samples of the language.
For example, if an English encoder is shared between En-De and En-Fr directions (and English sentences of two are not completely shared), the encoder learns more diverse English sentences from both pairs.
Few studies distinguished this effect \citep{firat2016multi,firat2016zero}.
We refer to the improvement resulting from this factor as the \textit{data-diversification} effect. 

\paragraph{Regularization}
Learning to encode or decode the same language in various directions may result in better representation learning and less overfitting in a single direction.
This effect has already been observed by \citet{firat2016multi} as the benefit of generalization and suggested by \citet{aharoni2019massively} to benefit many-to-many models compared to many-to-one models.

\section{Comparison of single models, 1-1 and M2}
\label{sec:comparison}

We compared the models with the same inference capacity in a series of conditions. 
Note that most of the multilingual NMT research was conducted in a joint one-to-many and many-to-one environment (JM2M): collected data are English-centered.
Despite its simplicity, observations under such setting may be unreliable to speak for many-to-many (M2M) environment, which is also clearly in demand in the industry. 
Therefore, we set M2M training as the default.

We also distinguish between two different dataset compositions: the \textit{sharing} case where all language pairs share the same sentence set, and the \textit{non-sharing} case where there is no overlap between different pairs.
To illustrate, a multiparallel set `En - Es - Ko' can be shared for all possible three pairs (En - Es, En - Ko, Es - Ko) or used only once for one pair.
Considering that multiparallel data is rare in practice, we compared the models in a strictly non-sharing environment.

\subsection{Settings}

\paragraph{Dataset}

We collected multi-parallel data from Europarl \cite{koehn2005europarl} and selected four languages: German, English, Finnish, and French. 
To construct a completely balanced environment, we created 500K, 10K, and 10K (train, valid, and test) non-sharing pairs for every twelve possible directions from 1.56M multi-parallel data. 
For the unbalanced environment, we synthetically reduced the amount of data for some pairs to match a specific ratio of the data amounts for low, medium, and high resource pairs.
For further details on data division, see appendix \ref{app:dataset}.

\paragraph{Model}
For the 1-1 model, we used the model of \citet{aharoni2019massively} which is transformer implementation of \citet{johnson2017google}. 
For the M2, we modified \citet{firat2016multi} to not share the attention module. 
Language-specific embeddings are shared between the encoder and decoder.
We implemented all models using transformer \citep{vaswani2017attention}.
We used the transformer with a hidden dimension of 256 and a feed-forward dimension of 1024 for our base model.
The rest of the configuration follows the base model employed by \citet{vaswani2017attention} except for the attention dropout and activation dropout of 0.1.
The 1-1 model uses a joint vocabulary with 32K tokens, whereas the M2 uses a language-specific vocabulary with 16K tokens each, all processed using the BPE \citep{kudo2018subword} of the \texttt{\small sentencepiece} package\footnote{\texttt{\small https://github.com/google/sentencepiece}} \citep{kudo2018sentencepiece}.

\paragraph{Training}
We used the \texttt{\small fairseq} framework \footnote{\texttt{\small https://github.com/pytorch/fairseq}} \citep{ott2019fairseq} to train and test all models. 
We set the batch size so that every encoder/decoder module learned at a maximum of 6144 tokens/GPU.
All models were trained using 4 NVIDIA Tesla V100 GPUs.
We followed the default parameters of the Adam optimizer \citep{kingma2014adam}.
For the learning rate schedule, we used 2K warm-up steps until 1e-3, after which we used the inverse square root learning rate schedule \citep{vaswani2017attention}.
The best model was selected using the best validation loss within the same maximum number of epochs.
All the performance was measured in \texttt{\small sacreBLEU4} \citep{post2018call} using a beam size of 4 and a length penalty of 0.6. 
Appendix \ref{app:training} provides more details of training.

\begin{table}
\centering
\begin{tabular}{|c|c|c|c|}
\hline
\textbf{Pairs} & \textbf{Single} & \textbf{1-1} & \textbf{M2} \\
\hline\hline\textbf{}
De-En &  33.00 &  31.04 (-1.96) &  33.51 (0.51) \\
De-Fi &  15.20 &  13.08 (-2.12) &  15.93 (0.73) \\
De-Fr &  28.47 &  25.73 (-2.74) &  29.08 (0.61) \\
En-De &  25.87 &  23.83 (-2.04) &  26.46 (0.59) \\
En-Fi &  19.57 &  16.94 (-2.63) &  20.03 (0.46) \\
En-Fr &  35.74 &  32.99 (-2.75) &  36.09 (0.35) \\
Fi-De &  18.97 &  16.75 (-2.22) &  19.51 (0.54) \\
Fi-En &  29.26 &  27.32 (-1.94) &  30.24 (0.98) \\
Fi-Fr &  25.21 &  22.24 (-2.97) &  25.94 (0.73) \\
Fr-De &  22.23 &  20.09 (-2.14) &  22.64 (0.41) \\
Fr-En &  35.49 &  33.81 (-1.68) &  36.18 (0.69) \\
Fr-Fi &  15.42 &   13.6 (-1.82) &  16.15 (0.73) \\
\hline
Avg  &  25.37 &  23.12 (-2.25) &  \textbf{25.98 (0.61)} \\
\hline
\end{tabular}
\caption{ SacreBLEU test scores of single models, 1-1, and M2 trained using a completely balanced, non-sharing dataset. Values in parentheses indicate the performance difference from single models. }
\label{table:bnc}
\end{table}

\subsection{Balanced environment}

\begin{table*}
\centering
\begin{tabular}{|c|c|c|c|c|c|c|}
\hline
\textbf{ID} & \textbf{Data sharing} & \textbf{Model size} & \textbf{Training pairs} & \textbf{Single} & \textbf{1-1} & \textbf{M2} \\
\hline\hline
1 & Non-sharing & Base & M2M(12) & 25.37 & 23.12 (-2.25) & \textbf{25.98 (0.61)}\\
\hline
2 & Sharing & Base & M2M(12) & 25.34 & 23.27 (-2.07)  & \textbf{25.65 (0.31)}\\
\hline
3 & Non-sharing & Large & M2M(12) & 25.43 & 26.90 (1.47) & \textbf{27.17 (1.74)} \\
\hline\hline
$4^*$ & Non-sharing & Base & JM2M(6) & - & 27.50 (-2.32) & 29.70 (-0.12) \\
\hline
$5^*$ & Non-sharing & Base & M2M(12) & 29.82 & 27.66 (-2.16) & \textbf{30.42 (0.6)} \\
\hline
\end{tabular}
\caption{ Averaged SacreBLEU test scores of single models, 1-1, and M2 trained using a balanced dataset of different configurations. \textit{M2M} indicates the training of full many-to-many directions among languages (12 directions), whereas \textit{JM2M} represents the training of directions that only include English on one side(6 directions). $^*$ indicates that the score is averaged only on English-centric.}
\label{table:bab}
\end{table*}

We first compared the performance of multi-way directions in a balanced and non-sharing environment, which is the most strictly controlled.

The results are shown in Table \ref{table:bnc}. 
The 1-1 model performed worse than both the single models and the M2 in every direction, clearly indicating a capacity bottleneck.
In contrast, \textbf{the M2 consistently outperformed not only the 1-1 model but also the single models in all directions}. 
As the M2 cannot benefit from \textit{cross-language effect} due to the lack of a shared module between any languages, we hypothesize that the following two effects are in charge: \textit{data-diversification} and \textit{regularization}.
We verify this hypothesis using ablation studies.

Note that the 1-1 model’s variation of degradation is higher with target languages than with source languages, even though all the directions are trained using the same amount.
The translation to English (-1.96, -1.94, and -1.68) consistently degraded the least, whereas that to French (-2.74, -2.75, and -2.97) degraded the most, given the same source languages.
This finding is consistent with previous observations that the capacity bottleneck is more prominent in the decoders \citep{johnson2017google,arivazhagan2019massively}. 

\label{subsec:unbalanced}

\begin{table*}
\centering
\small
\begin{tabular}{|c|c|c|c|c|c|c|c|}
\hline
\multirow{2}{*}{\textbf{Resource}} & \multirow{2}{*}{\textbf{Pairs}} &\multicolumn{2}{c|}{\textbf{1:1:1}}  &\multicolumn{2}{c|}{\textbf{1:2:4}}  &\multicolumn{2}{c|}{\textbf{1:5:25}} \\
\cline{3-8}
& & \textbf{1-1} &   \textbf{M2} & \textbf{1-1} & \textbf{M2} & \textbf{1-1} & \textbf{M2} \\ 
\hline\hline
\multirow{3}{*}{High} & En-Fi      &  16.94 (-2.63) &  20.03 (0.46) &    18.01 (-1.4) &   19.92 (0.51) &   18.66 (-0.75) &   19.82 (0.41) \\
& Fi-En      &  27.32 (-1.94) &  30.24 (0.98) &   28.04 (-1.21) &   30.06 (0.81) &   28.51 (-0.74) &    29.9 (0.65) \\
\cline{2-8}
& Avg    &  22.13 (-2.28) &  25.14 (0.72) &    23.02 (-1.3) &   \textbf{24.99 (0.66)} &   23.58 (-0.74) &   \textbf{24.86 (0.53)} \\
\hline
\multirow{5}{*}{Medium} & En-Fr      &  32.99 (-2.75) &  36.09 (0.35) &   32.73 (-1.24) &   35.26 (1.29) &    31.61 (1.14) &   33.66 (3.19) \\
& Fr-En      &  33.81 (-1.68) &  36.18 (0.69) &   33.73 (-0.23) &   35.39 (1.43) &      33.1 (2.5) &     33.9 (3.3) \\
& Fi-Fr      &  22.24 (-2.97) &  25.94 (0.73) &   23.35 (-0.11) &   25.27 (1.81) &     22.6 (3.37) &   24.08 (4.85) \\
& Fr-Fi      &   13.6 (-1.82) &  16.15 (0.73) &    14.49 (0.47) &   15.58 (1.56) &    14.43 (3.78) &   14.19 (3.54) \\
\cline{2-8}
& Avg    &  25.66 (-2.31) &  28.59 (0.62) &   26.08 (-0.28) &   \textbf{27.88 (1.52)} &     25.44 (2.7) &   \textbf{26.46 (3.72)} \\
\hline
\multirow{7}{*}{Low} & De-En      &  31.04 (-1.96) &  33.51 (0.51) &    30.31 (1.68) &   32.29 (3.66) &   28.45 (17.02) &  27.88 (16.45) \\
& En-De      &  23.83 (-2.04) &  26.46 (0.59) &     22.69 (0.9) &   24.78 (2.99) &   18.61 (11.66) &  19.91 (12.96) \\
& De-Fi      &  13.08 (-2.12) &  15.93 (0.73) &    13.72 (2.56) &   14.89 (3.73) &   12.76 (10.58) &   11.62 (9.44) \\
& Fi-De      &  16.75 (-2.22) &  19.51 (0.54) &     16.8 (2.06) &    18.3 (3.56) &   14.01 (10.99) &  14.25 (11.23) \\
& De-Fr      &  25.73 (-2.74) &  29.08 (0.61) &     25.8 (1.45) &    27.6 (3.25) &   23.37 (15.76) &   23.5 (15.89) \\
& Fr-De      &  20.09 (-2.14) &  22.64 (0.41) &    19.76 (1.35) &   21.45 (3.04) &   16.18 (10.76) &  16.58 (11.16) \\
\cline{2-8}
& Avg    &   21.75 (-2.2) &  24.52 (0.57) &    21.51 (1.67) &   \textbf{23.22 (3.37)} &     18.9 (12.8) &  \textbf{18.96 (12.85)} \\
\hline
& Total Avg &  23.12 (-2.25) &  25.98 (0.61) &    23.29 (0.52) &    \textbf{25.07 (2.3)} &    21.86 (7.17) &   \textbf{22.44 (7.76)} \\
\hline
\end{tabular}
\caption{ Test SacreBLEU test scores of single models, 1-1 model, and M2 trained using an unbalanced, completely non-sharing dataset. 1:1:1, 1:2:4, and 1:5:25 represent the ratios of the low, medium, and high resource pairs, respectively. Values in parentheses indicate the performance difference from single models in respective environments. }
\label{table:ubnc}
\end{table*}

\paragraph{Ablation}

We compare models in a series of conditions (see IDs in Table \ref{table:bab}).
\textcircled{\raisebox{-.9pt}1} We denote the summarized performance demonstrated in Table \ref{table:bnc} for reference.
\textcircled{\raisebox{-.9pt}2} To establish whether \textit{data-diversification} was responsible for the performance improvement of the M2, we experimented using fully shared data.
\textcircled{\raisebox{-.9pt}3} To observe the behavior under alleviated capacity constraints, we experimented using bigger models. 
We used a transformer with a hidden dimension of 512 and a feed-forward dimension of 2048 for our large model. 
The training settings are the same except for a larger batch size (x4).
\textcircled{\raisebox{-.9pt}4} Finally, we compared the models trained using the JM2M (6 directions instead of 12) to observe the behavior of the models with fewer directions.
\textcircled{\raisebox{-.9pt}5} We averaged scores of English-centric directions in \textcircled{\raisebox{-.9pt}1} to compare with \textcircled{\raisebox{-.9pt}4}.
Appendix \ref{app:detail} presents the individual score for each direction.

Table \ref{table:bab} shows the results of each environment.
When we completely shared the data(\textcircled{\raisebox{-.9pt}2}), the performance gain of the M2 versus that of the single models (0.31) decreased.
Given that \textcircled{\raisebox{-.9pt}2} eliminates the chance of \textit{data-diversification}, the degraded performance (0.3) can be attributed to it.
However, the fact that the M2 still outperforms the single models (0.31) implies that the M2 can still benefit from the \textit{regularization} effect of multi-way training.
The minor increase in performance of 1-1 (0.18) seems to imply that \textit{data-diversification} can be detrimental under the severe capacity bottleneck. 

\textcircled{\raisebox{-.9pt}3} shows the performance of a larger model trained using the same data.
Single models barely improved with the use of larger models, indicating the absence of a capacity bottleneck.
On the contrary, the 1-1 model and the M2 both showed an increase in performance.
The 1-1 model exhibits a gain from multi-way training only with enough capacity (1.47).
This indicates that the benefit of multi-way training can only be achieved with enough capacity for the 1-1 model.
Although the M2 is less affected by capacity bottleneck, the larger capacity is also beneficial for the M2 (1.74) to fully leverage the benefits of multi-way training.

To compare the models trained with JM2M (\textcircled{\raisebox{-.9pt}4}), \textcircled{\raisebox{-.9pt}5} shows the score averaged only over directions from and to English \textcircled{\raisebox{-.9pt}1}.
The JM2M scheme is likely to have mixed results: there is less pressure from the capacity bottleneck due to fewer training directions. 
However, possible gains from \textit{data-diversification} or \textit{regularization} are also smaller.
Both the 1-1 model and the M2 perform better when trained using M2M (\textcircled{\raisebox{-.9pt}5}) than when trained using JM2M (\textcircled{\raisebox{-.9pt}4}).
However, the performance difference is more significant in the M2 (0.72) than in the 1-1 model (0.16).
We assume that while both models benefit from \textit{data-diversification} and \textit{regularization} accompanied by training using more directions, the capacity bottleneck in 1-1 counterweighs those positive effects.  

\subsection{Unbalanced environment}
\label{subsec:unbalanced}

We also compared the models with unbalanced training data, which is a natural condition in practice.
To synthetically create an unbalanced environment, we first divided the pairs into low (De-En, De-Fi, De-Fr), medium (En-Fr, Fi-Fr), and high (En-Fr) resource pairs.
Next, we reduced the amount of data for low and medium pairs, setting the ratio of low:medium:high = 1:2:4, and 1:5:25, respectively. 
The detailed division of the dataset can be found in appendix \ref{app:dataset}.
Note that the models learns with fewer data in the unbalanced environment.
We first trained the models without up-sampling. 

Table \ref{table:ubnc} shows the scores of the 1-1 model and the M2 in each setting (1:1:1, 1:2:4, 1:5:25). 
Both models show similar trends with unbalanced data.
Compared to the balanced environment, medium and low resource pairs tend to benefit from multi-way training, with gains more prominent for lower resource pairs as the data get more unbalanced (12.8 by the 1-1 model and 12.85 by the M2).
Interestingly, \textbf{the M2 exhibits a similar level of improvement to that of the 1-1 model in low and medium resource pairs.}
Considering the M2 is not subject to the \textit{cross-language transfer}, the performance increase in lower resource pairs may be better explained by \textit{data-diversification} and \textit{regularization}.
This indicates that the \textit{cross-language effect} of the 1-1 model may be more subtle than expected.

On the other hand, M2 barely showed the performance degradation in high resource pairs.
This implies that the performance boost of low resource pairs and the drop of high resource pairs may not be necessarily trade-off without a capacity bottleneck. 

\paragraph{Ablation}
The sampling method in an unbalanced setting is known to affect the performance \citep{arivazhagan2019massively}.
We compared two models in the most unbalanced environment (1:5:25) with and without up-sampling.

\begin{table}
\centering
\small
\begin{tabular}{|c|c|c|c|c|}
\hline
\textbf{M} & \textbf{US} & \textbf{High} & \textbf{Medium} & \textbf{Low} \\
\hline\hline
\multirow{2}{*}{1-1} & $\times$ & 23.58 (-0.74) & 25.44 (2.7) & 18.9 (12.8) \\
\cline{2-5}
& $\circ$ & 20.5 (-3.83) & 24.31 (1.57) & \textbf{19.78 (13.68)} \\
\hline
\multirow{2}{*}{M2} & $\times$ & \textbf{24.86 (0.53)} & \textbf{26.46 (3.72)} & 18.96 (12.85) \\
\cline{2-5}
& $\circ$ & 19.64 (-4.69) & 23.49 (0.75) & 16.88 (10.78) \\
\hline
\end{tabular}
\caption{ Averaged test SacreBLEU scores of 1-1 and M2 trained with 1:5:25 dataset with and without up-sampling. }
\label{table:ubnusc}
\end{table}

Table \ref{table:ubnusc} shows the results.
As previously reported, we confirm that up-sampling makes the results extreme in the 1-1 model: low resource pairs improve more (from 12.8 to 13.68), whereas high resource pairs degrade more (from -0.74 to -3.83).
On the other hand, up-sampling in the M2 harmed performance in all the low, medium, and high resource pairs.
The difference in converge rates among modules may be the cause; models overfit in low-resource pairs, and underfit in high-resource pairs. 
This is supported by the changes in the M2’s performance with more training epochs (Appendix \ref{app:detail}).

\begin{figure}[h]
  \centering
  \includegraphics[width=0.45\textwidth]{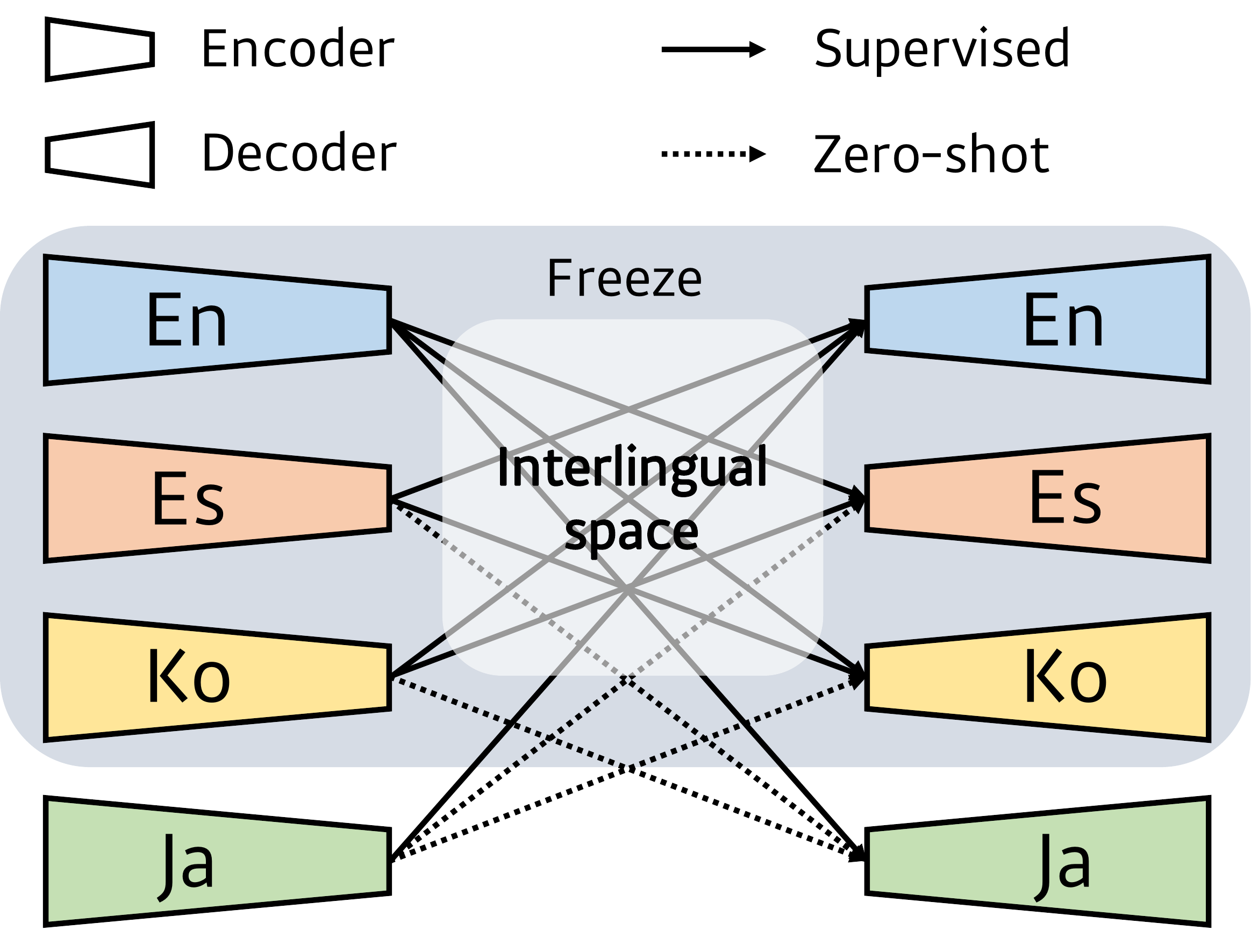}
  \caption{\textbf{Interlingual space} formed by the multi-way training of the M2 (En, Es and Ko). While freezing the M2, incrementally training a new language (Ja) with a single parallel corpus (En - Ja) adapt new modules to the interlingual space. }
  \label{fig:space}
\end{figure}

\section{Interlingual space of the M2}
\label{sec:interlingual}

Creating interlingual space has been an active research area \citep{lu2018neural, sen2019multilingual, arivazhagan2019missing, escolano2019bilingual, vazquez2019multilingual} because it is critical to scaling out languages, such as incremental learning.
Because input of M2 does not contain any information regarding the target language, encoders need to encode it so that any decoder can translate.
At the same time, decoders of the M2 should be able to generate from output of any M2 encoder.
For this reason, we assume that the output space of M2 encoders is interlingual.

Figure \ref{fig:space} illustrates the interlingual space of a M2. Multi-way training of 3 languages (En, Es and Ko) forms the interlingual space which is shared by 6 modules. 
This space is preserved as long as the weights of the M2 are frozen. 
Training a new module (Ja) with a single parallel corpus (En - Ja) using one of the frozen modules (En) adapt the module to the interlingual space. 
We speculate that the new module (Ja) would be compatible with the other modules (Es and Ko) if the interlingual space is formed well. 

We verify this using incremental zero-shot learning.
Additionally, we measure how the language invariance of the space changes as the number of languages involved in the M2 varies.
Since maintainability is one of the critical needs in practice, high performance on incremental learning would be a desirable trait in industrial settings.

\subsection{Setting}

To increase the number of languages, we modified the multi-parallel corpus of Europarl differently. 
We selected six languages (German, English, Spanish, Finnish, French, and Dutch) and divided a 1.25M multi-parallel corpus into 250K for each direction without sharing.
Other details are mostly the same as in former experiments.
The detailed division of the dataset and training details can be found in appendix \ref{app:dataset} and \ref{app:training}.

\begin{table}
\centering
\begin{tabular}{|c|l|cc|}
\hline
\textbf{ID} &\textbf{Model} &  \textbf{En-Fr} &  \textbf{Fr-En} \\
\hline\hline
1 & M2(4) + En-Fr &  34.70 &  34.90 \\
2 & M2(4) + En-Fr \textit{with init} &  34.88 &  34.94 \\
3 & M2(4) + En,De-Fr &  35.40 &  35.57 \\
4 & M2(4) + En,De,Es-Fr &  35.41 &  35.70 \\
5 & M2(4) + En,De,Es,Nl-Fr &  \textbf{35.47} &  \textbf{35.92} \\
\hline
$6^*$ & Single &  34.48 &  34.11 \\
$7^*$ & M2(5) &  36.24 &  36.35 \\
\hline
\end{tabular}
\caption{ SacreBLEU test scores of a single model and incremented modules of the M2. Values in parentheses indicate the number of languages involved in the M2 (4: De, En, Es, Nl; 5: 4 + Fr). + indicates incremental training with the former model frozen. \textit{with init} indicates that the incremented module is initialized using the weight of the English module. $^*$ represents the model is trained from scratch and not incrementally. }
\label{table:inc}
\end{table}

\subsection{Incremental training}

\begin{table*}
\centering
\begin{tabular}{|c|cccccc|}
\hline
\textbf{Model} &  \textbf{De-Fr} &  \textbf{Fr-De} &  \textbf{Es-Fr} &  \textbf{Fr-Es} &  \textbf{Nl-Fr} &  \textbf{Fr-Nl} \\
\hline\hline
Pivot &  25.42 &  19.53 &  30.37 &  30.87 &  23.52 &  22.06 \\
1 &  26.37 &  19.08 &  31.91 &  32.22 &  24.31 &  22.15 \\
2 &  \textbf{26.79} &  \textbf{19.90} &  32.17 &  32.68 &  24.64 &  22.63 \\
3 &  - &  - &  \textbf{32.91} &  \textbf{33.34} &  25.65 &  23.44 \\
4 &  - &  - &  - &  - &  \textbf{25.82} &  \textbf{23.55} \\
\hline
$6^*$ & 26.91 & 20.86 & 32.90 & 33.70 & 24.81 & 22.97 \\
$7^*$ &  28.86 &  22.70 &  34.62 &  35.22 &  26.58 &  24.98 \\
\hline
\end{tabular}
\caption{ SacreBLEU zero-shot test scores of the English-pivoted single models and incremented modules from Table \ref{table:inc}. $^*$ means that the model is trained using the supervision of 250 thousand pairs. }
\label{table:zer}
\end{table*}

We added French to an M2 model trained using all directions among four languages (German, English, Spanish, and Dutch).
An additional French encoder and decoder were trained using English-French pairs while the parameters of English modules remained frozen (\textcircled{\raisebox{-.9pt}1}). 
We also tested two methods to help incremental training as follows.
1) Initialize the new module using one of the modules trained using other languages.
In the experiment, we used the weights copied from the English module as the initialization for French (\textcircled{\raisebox{-.9pt}2}).
Note that the English and French module does not share any information, such as embedding.
2) Train the module with auxiliary directions. 
We incrementally added auxiliary directions of De-Fr (\textcircled{\raisebox{-.9pt}3}), Es-Fr (\textcircled{\raisebox{-.9pt}4}), and Nl-Fr (\textcircled{\raisebox{-.9pt}5}).
We compared the models with a singly trained model (\textcircled{\raisebox{-.9pt}6}), and the M2 models trained using five languages from scratch (\textcircled{\raisebox{-.9pt}7}).
\textcircled{\raisebox{-.9pt}7} worked as an upper bound for the incremental training.

Table \ref{table:inc} shows the performance of En-Fr and Fr-En with incremental training.
\textbf{The incrementally trained model without any additional method (\textcircled{\raisebox{-.9pt}1}) outperformed a single model (\textcircled{\raisebox{-.9pt}6}) even though half of the model was frozen.}
This not only indicates that the language-agnostic space is well-formed but also shows that incremented direction can benefit from a well-trained frozen module.

We also found that our two methods are effective in incremental training.
Even though French does not share any information with the trained English module, initializing the French module with the weights learned by the English module benefits the performance marginally.
Incrementally training the new module using multiple directions helps as the number of directions increases.
Note that the two methods can be applied orthogonally.
Although none of the incrementally trained models outperform the M2 model trained from scratch, this still shows that simple incremental training for the M2 can be a good alternative for expensive training from scratch.

We examined whether an incremented module in one direction can generalize to the other directions.
We compared the zero-shot performance of the models in Table \ref{table:inc} with the English-pivoted translation performance using two single models. 
We also denoted the supervised performance of single models, and jointly trained the M2 for reference (250K for each direction).

\subsection{Incremental zero-shot learning}

Table \ref{table:zer} shows the zero-shot performance of incrementally trained modules. 
Amazingly, \textbf{most of the incremented modules demonstrated better performance than the English-pivoted translation}.
The only exception was in the Fr-De direction of the naively incremented module (\textcircled{\raisebox{-.9pt}1}), which seemed to be marginal (-0.45).
Our methods for incremental training were also effective for zero-shot performance.
The results were even comparable to the single supervised models trained with 250K parallel corpus.
This shows that multi-way training creates shared (interlingual) space instead of pair specific space.

\subsection{The language invariance of the interlingual space}

\begin{table*}
\centering
\begin{tabular}{|c|ccc|c||c|}
\hline
\multirow{2}{*}{\textbf{Model}} & \multicolumn{4}{c||}{\textbf{Cosine Similarity}} & \textbf{BLEU Score} \\
\cline{2-6}
&  \textbf{En-De} &  \textbf{En-Nl} &  \textbf{De-Nl} & \textbf{Avg} & \textbf{En-En}\\
\hline\hline
M2(3) & 0.7228 &  0.7062 &  0.7043 &  0.7111 & 75.55 \\
M2(4) & 0.7682 &  0.7425 &  0.7635 &  0.7581 & 82.55 \\
M2(5) & 0.7832 &  0.7603 &  0.7827 &  0.7754 & \textbf{83.13} \\
M2(6) & 0.8169 &  0.7905 &  0.8189 &  \textbf{0.8088} & 82.80 \\
\hline
\end{tabular}
\caption{ Cosine text similarity score of encoder outputs and SacreBLEU score of mono-direction translation(En-En). Values in parentheses indicate the number of languages involved in the M2 (3: De, En, Nl; 4: 3 + Es; 5: 4 + Nl; 6: 5 + Fi). }
\label{table:qs}
\end{table*}

The interlingual space established by the M2 was confounding, considering no additional regularizations or methods were adopted.
We measured the language invariance of the interlingual space while the varying the number of languages of the M2 model.
We trained a series of M2 models that included 3 - 6 languages (6, 12, 20, and 30 directions) and found that the use of more languages to train the M2 also improved its performance in all directions (appendix \ref{app:languages}).
We investigated with two metrics to measure the language invariance of interlingual space.

\paragraph{Cosine Similarity}
We measured the representation similarity of parallel sentences from a parallel corpus. 
To obtain the fixed-size representation, we average pooled the output of encoders through the time steps.
We averaged the cosine similarity of 10K pairs from the test set.

\paragraph{Mono-direction translation}
When training the M2, mono-direction (where source and target languages are the same) is not trained because modules tend to learn to simply copy the input, which hinders translation training \citep{firat2016multi}. 
Meanwhile, interlingual output representation of the encoders should be able to be translated by any decoder, including the decoder of the source language.
Therefore, the translation score of mono-direction translation shows how well the information of the source sentence is preserved.

Table \ref{table:qs} shows the cosine similarity and mono-direction translation scores of the M2.
As the M2 trains using more languages, the cosine similarity of all three pairs increases, which implies higher language invariance in interlingual space.
However, the gain from marginal languages decreases as the number of languages increases.
Mono-direction translation scores mostly align with the number of languages except for the M2(6), which degraded a little from M2(5).
As a result, we reasonably conclude that the language invariance of the interlingual space improves with more languages.


\section{Conclusion}
In this study, we re-evaluate the M2 model and suggest it as an appropriate choice for multilingual translation in industries.
By extensively comparing the single models, 1-1 model, and M2 in varying conditions, we find that the M2 can benefit from multi-way training through \textit{data-diversification} and \textit{regularization} while suffering less from capacity bottlenecks.
Additionally, we demonstrate that the M2 can also benefit low resource pairs in an unbalanced environment as a 1-1 model without being subject to \textit{cross-language effect}.
Next, we suggest that the M2 model is easily maintainable because of its interlingual space.
The interlingual space not only enables incremental training in a simple manner, but also accompanies competitive incremental zero-shot performance. 
Furthermore, we validate that the language invariance of the space enhances as the number of languages in the M2 increases.
We hope that this study sheds light on the relatively disregarded M2 model and provide a benchmark for selecting a model among varying levels of shared components.


\section*{Acknowledgments}
The authors of this paper would like to give special thanks to Sang-Woo Lee and Jihyung Moon for their honest and genuine feedback that helped improve the quality of the paper greatly.
Also, we are extremely grateful to Jaehyeon Kim, Jaehun Jung and the anonymous reviewers for their suggestions and comments.
Finally, we cannot express enough thanks to all of Context Part members of Kakao Enterprise for their continued support and encouragement. 

\bibliography{emnlp2020}
\bibliographystyle{acl_natbib}

\appendix

\section{Dataset}
\label{app:dataset}

\subsection{Division of multi-parallel dataset}

\begin{table}
\centering
\begin{tabular}{|c||c|c|c|c|}
\hline
& De & En & Fi & Fr \\
\hline\hline
De & - & 1 & 2 & 3 \\
\hline
En & - & - & 3 & 2 \\
\hline
Fi & - & - & - & 1 \\
\hline
Fr & - & - & - & - \\
\hline
\end{tabular}
\caption{ Division of multi-parallel parts for each pair in section \ref{sec:comparison} }
\label{table:div1}
\end{table}

\begin{table}
\centering
\begin{tabular}{|c||c|c|c|c|c|c|}
\hline
& De & En & Es & Fi & Fr & Nl \\
\hline\hline
De & - & 1 & 2 & 3 & 4 & 5 \\
\hline
En & - & - & 3 & 4 & 5 & 2 \\
\hline
Es & - & - & - & 5 & 1 & 4 \\
\hline
Fi & - & - & - & - & 2 & 1 \\
\hline
Fr & - & - & - & - & - & 3 \\
\hline
Nl & - & - & - & - & - & - \\
\hline
\end{tabular}
\caption{ Division of multi-parallel parts for each pair in section \ref{sec:interlingual}}
\label{table:div2}
\end{table}

In order to create completely non-sharing dataset and make the best use of multi-parallel corpus, we divide the 1.5K multi-parallel corpus into 3 parts(500K) for section \ref{sec:comparison} and 5 parts(250K) for section \ref{sec:interlingual}. 
And then, we assigned the parts to pairs so that no two directions of the same side share the same part. 
The assignment for section \ref{sec:comparison} and \ref{sec:interlingual} are stated in table \ref{table:div1} and \ref{table:div2} respectively. 
Validation and test are divided with the same manner.
For complete-sharing dataset, training data for all pairs only created from part 1. 
However, validation and test set remain the same with completely non-sharing dataset.

\subsection{Amount of data for each pairs}
\label{sec:amount}

\begin{table}
\centering
\begin{tabular}{|c|c|c|c|c|}
\hline
\textbf{Resource} & \textbf{Pairs} & \textbf{1:1:1} & \textbf{1:2:4} & \textbf{1:5:25} \\
\hline\hline
High & En-Fi & 500K & 500K & 500K \\
\hline
\multirow{2}{*}{Medium} & En-Fr & 500K & 250K & 100K \\
& Fi-Fr & 500K & 250K & 100K \\
\hline
\multirow{3}{*}{Low} & De-En & 500K & 125K & 20K \\
& De-Fi & 500K & 125K & 20K \\
& De-Fr & 500K & 125K & 20K \\
\hline
\end{tabular}
\caption{ The amount of data for each pair in section \ref{sec:comparison} }
\label{table:amount}
\end{table}

In order to create unbalanced environment in section 4, we limited the amount of data for some directions. 
Table \ref{table:amount} shows the amount of the data for each pair in balanced, and unbalanced environments in section \ref{sec:comparison}. 
For section \ref{sec:interlingual}, the amounts of all directions are the same with 250K.
All the validation and test set are the same with 10K.

\bigbreak

Though our dataset can easily be reconstructed from the open dataset (Europarl) with described process, we also made our dataset available online\footnote{\texttt{\small https://drive.google.com/file/d/\\1CmSzFI6h2cGYJshUWEPkF7Hx4UcL3DVl}} for convenience of readers. 
We only uploaded the dataset of the balanced environment since unbalanced environment can be made from them trivially. 
The dataset is binarized with \texttt{\small fairseq-preprocess} command of \texttt{\small fairseq} framework.

\section{Training detail}
\label{app:training}

\subsection{Batch size}
We selected the batch size of 6144 max tokens with the best validation loss of a single model (En-De) among \{1536, 3072, 6144, 12288, 24576\} max tokens per GPU (4 GPUs).
While the total number of parameters and the training directions is different among single model, 1-1 and M2, we set the batch size for each direction so that each module learns with the same batch size (6144 tokens). 
Specifically, one step of a single model includes a single direction, while that of 1-1(4) and M2(4) includes 12 directions.
However, training directions per module between 1-1(4) and M2(4) is different with 12 and 3 directions.
Therefore, the batch size per direction of 1-1 is 512 (1/12 of 6144) and that of M2 is 1536 (1/4 of 6144).
Since we accumulate the gradients of all directions, all the compared modules learn with the same batch size of data.  

\subsection{Sampling}
To train balanced data, we used round robin scheduling of all directions.
We compared two sampling methods in ablation of unbalanced environment: up-sampling and proportional sampling. 
Round robin scheduling is equivalent to up-sampling low-resource data in unbalanced environment. 
For efficient proportional sampling, we sampled several small batches of pairs proportional to the amount of total pairs. 
We accumulated gradients of several batches to make expected batch-size of each module to meet the total batch size.

\subsection{Early stopping}
Since fixing the maximum tokens of a batch per module results in different step size among models, we stopped the training of models based on the maximum number of epochs.
All the best models were chosen based on the best validation loss (averaged) within 100 epochs.

\section{Detailed scores of ablations}
\label{app:detail}

This section provides detailed scores of the ablation part of the section \ref{sec:comparison} and \ref{sec:interlingual}.

\begin{table*}
\centering
\small
\begin{tabular}{|c|c|c|c|c|c|c|}
\hline
\multirow{2}{*}{\textbf{Pairs}} & \multicolumn{3}{c|}{\textbf{Sharing}} & \multicolumn{3}{c|}{\textbf{Large}} \\
\cline{2-7}
& \textbf{Single} & \textbf{1-1} & \textbf{M2}& \textbf{Single} & \textbf{1-1} & \textbf{M2} \\
\hline\hline
De-En &   32.86 &  30.61 (-2.25) &  33.14 (0.28) &   32.84 &   34.5 (1.66) &  34.59 (1.75) \\
De-Fi &   15.04 &  13.67 (-1.37) &  15.45 (0.41) &   15.21 &  16.96 (1.75) &  17.15 (1.94 \\
De-Fr &   28.27 &  26.39 (-1.88) &  28.93 (0.66) &   28.52 &  29.84 (1.32) &  30.24 (1.72) \\
En-De &   25.98 &  23.57 (-2.41) &  26.15 (0.17) &   25.97 &  27.23 (1.26) &   27.57 (1.6) \\
En-Fi &   19.52 &  17.19 (-2.33) &  19.63 (0.11) &   19.49 &  21.26 (1.77) &   21.39 (1.9) \\
En-Fr &   35.51 &  32.48 (-3.03) &  35.65 (0.14) &   35.65 &  36.39 (0.74) &  37.14 (1.49) \\
Fi-De &   18.63 &   17.33 (-1.3) &  19.38 (0.75) &   19.07 &  20.55 (1.48) &  20.88 (1.81) \\
Fi-En &   29.25 &  27.09 (-2.16) &  29.89 (0.64) &   29.39 &  31.35 (1.96) &  31.57 (2.18) \\
Fi-Fr &   25.53 &  23.18 (-2.35) &  25.78 (0.25) &   25.45 &  26.84 (1.39) &  27.39 (1.94) \\
Fr-De &   22.18 &   20.58 (-1.6) &  22.35 (0.17) &   22.28 &  23.73 (1.45) &  23.85 (1.57) \\
Fr-En &   35.58 &   33.2 (-2.38) &  35.63 (0.05) &   35.78 &  36.89 (1.11) &  37.12 (1.34) \\
Fr-Fi &   15.70 &  13.99 (-1.71) &  15.85 (0.15) &   15.55 &  17.28 (1.73) &  17.17 (1.62) \\
\hline
Avg  &   25.34 &  23.27 (-2.06) &  25.65 (0.32)  &   25.43 &   26.9 (1.47) &  27.17 (1.74) \\
\hline
\end{tabular}
\caption{ Detailed scores of \textcircled{\raisebox{-.9pt}2} and \textcircled{\raisebox{-.9pt}3} in table \ref{table:bab} }
\label{table:hnl}
\end{table*}

\begin{table*}
\centering
\small
\begin{tabular}{|c|c|c|c|c|c|}
\hline
\multirow{2}{*}{\textbf{Pairs}} & \multirow{2}{*}{\textbf{Single}} & \multicolumn{2}{c|}{\textbf{JM2M}} & \multicolumn{2}{c|}{\textbf{M2M}} \\
\cline{3-6}
&& \textbf{1-1} & \textbf{M2} & \textbf{1-1} & \textbf{M2} \\
\hline\hline
De-En        &   33.00 &  30.93 (-2.07) &  32.55 (-0.45) &  31.04 (-1.96) &  33.51 (0.51) \\
Fi-En        &   29.26 &  27.18 (-2.08) &  29.08 (-0.18) &  27.32 (-1.94) &  30.24 (0.98) \\
Fr-En        &   35.49 &  33.84 (-1.65) &    35.6 (0.11) &  33.81 (-1.68) &  36.18 (0.69) \\
En-De        &   25.87 &   23.6 (-2.27) &    25.9 (0.03) &  23.83 (-2.04) &  26.46 (0.59) \\
En-Fi        &   19.57 &   16.6 (-2.97) &  19.32 (-0.25) &  16.94 (-2.63) &  20.03 (0.46) \\
En-Fr        &   35.74 &  32.86 (-2.88) &   35.77 (0.03) &  32.99 (-2.75) &  36.09 (0.35) \\
\hline
Avg         &   29.82 &   27.5 (-2.32) &   29.7 (-0.12) &  27.66 (-2.17) &   30.42 (0.6) \\
\hline

\end{tabular}
\caption{ Detailed scores of \textcircled{\raisebox{-.9pt}4} and \textcircled{\raisebox{-.9pt}5} in table \ref{table:bab} }
\label{table:joint}
\end{table*}

Table \ref{table:hnl} shows detailed scores under complete sharing (\textcircled{\raisebox{-.9pt}2} of table \ref{table:bab}) and increased capacity (\textcircled{\raisebox{-.9pt}3} of table \ref{table:bab}).
Table \ref{table:joint} shows detailed scores under JM2M(\textcircled{\raisebox{-.9pt}4} of table \ref{table:bab}) and M2M(\textcircled{\raisebox{-.9pt}5} of table \ref{table:bab}) training.

\begin{table*}
\centering
\small
\begin{tabular}{|c|c|c|c|c|c|c|}
\hline
\multirow{2}{*}{\textbf{Resource}} & \multirow{2}{*}{\textbf{Pairs}} &\multicolumn{2}{c|}{\textbf{Proportional sampling}}  &\multicolumn{3}{c|}{\textbf{Up-sampling}} \\
\cline{3-7}
& & \textbf{1-1} &   \textbf{M2} & \textbf{1-1} & \textbf{M2} & \textbf{M2(+10)} \\ 
\hline\hline
\multirow{3}{*}{High} & En-Fi      &  18.66 (-0.75) &   19.82 (0.41) &   15.7 (-3.71) &  14.76 (-4.65) &  16.86 (-2.55) \\
&Fi-En      &  28.51 (-0.74) &    29.9 (0.65) &   25.3 (-3.95) &  24.52 (-4.73) &  26.81 (-2.44) \\
\cline{2-7}
&Avg    &  23.58 (-0.74) &   24.86 (0.53) &   20.5 (-3.83) &  19.64 (-4.69) &   21.84 (-2.5) \\
\hline
\multirow{5}{*}{Medium} & En-Fr      &   31.61 (1.14) &   33.66 (3.19) &    31.27 (0.8) &   30.79 (0.32) &   32.72 (2.25) \\
&Fr-En      &     33.1 (2.5) &     33.9 (3.3) &   31.75 (1.15) &   30.98 (0.38) &   32.39 (1.79) \\
&Fi-Fr      &    22.6 (3.37) &   24.08 (4.85) &   21.54 (2.31) &   20.64 (1.41) &    22.43 (3.2) \\
&Fr-Fi      &   14.43 (3.78) &   14.19 (3.54) &   12.67 (2.02) &   11.54 (0.89) &   12.86 (2.21) \\
\cline{2-7}
&Avg    &    25.44 (2.7) &   26.46 (3.72) &   24.31 (1.57) &   23.49 (0.75) &    25.1 (2.36) \\
\hline
\multirow{7}{*}{Low} & Ee-En      &  28.45 (17.02) &  27.88 (16.45) &  28.31 (16.88) &   24.5 (13.07) &  24.24 (12.81) \\
&En-De      &  18.61 (11.66) &  19.91 (12.96) &  21.03 (14.08) &  18.61 (11.66) &  18.32 (11.37) \\
&Ee-Fi      &  12.76 (10.58) &   11.62 (9.44) &    11.8 (9.62) &    9.21 (7.03) &    9.31 (7.13) \\
&Fi-De      &  14.01 (10.99) &  14.25 (11.23) &  15.06 (12.04) &    12.6 (9.58) &   12.27 (9.25) \\
&De-Fr      &  23.37 (15.76) &   23.5 (15.89) &  24.34 (16.73) &   20.81 (13.2) &  20.27 (12.66) \\
&Fr-De      &  16.18 (10.76) &  16.58 (11.16) &   18.12 (12.7) &  15.56 (10.14) &   14.94 (9.52) \\
\cline{2-7}
&Avg    &    18.9 (12.8) &  18.96 (12.85) &  19.78 (13.68) &  16.88 (10.78) &  16.56 (10.46) \\
\hline
& Total Avg &   21.86 (7.17) &   22.44 (7.76) &   21.41 (6.72) &   19.54 (4.86) &    20.28 (5.6) \\
\hline
\end{tabular}
\caption{ Detailed scores of models of table \ref{table:ubnusc}. M2(+10) indicates the selected best model trained with addtional 10 epochs. }
\label{table:us}
\end{table*}

Table \ref{table:us} shows detailed scores of the models under proportional sampling and up-sampling in table \ref{table:ubnusc}. 
M2(+10) indictes the scores of the M2 trained 10 epochs after the best validation loss. 
M2(+10) shows the increased performance in medium and high resource pairs and degradation in low resource pairs.
This indicates that up-sampling causes the difference in converge rates among pairs of different resources for M2.

\section{Detailed scores of M2 with varying languages}
\label{app:languages}

\begin{table}
\centering
\small
\begin{tabular}{|c|c|c|c|c|}
\hline
\textbf{Pair} & \textbf{M2(3)} & \textbf{M2(4)} & \textbf{M2(5)} & \textbf{M2(6)} \\
\hline\hline
De-En &     32.33 &        32.96 &           33.20 &              33.53 \\
En-De &     25.52 &        25.75 &           26.16 &              26.15 \\
De-Nl &     25.10 &        25.49 &           25.34 &              25.60 \\
Nl-De &     21.32 &        21.56 &           21.55 &              21.71 \\
En-Nl &     27.17 &        27.39 &           27.65 &              27.77 \\
Nl-En &     29.53 &        29.94 &           30.27 &              30.43 \\
\hline
Avg  &     26.83 &        27.18 &           27.36 &              27.53 \\
\hline
\end{tabular}
\caption{ Detailed scores of the models in \ref{table:qs} }
\label{table:languages}
\end{table}

Table \ref{table:languages} shows detailed scores of M2 trained with varying number of languages.
This shows that M2 trained with more languages shows better performance.

\end{document}